\pdfoutput=1
\documentclass[runningheads]{llncs}
\usepackage[T1]{fontenc}

\usepackage{graphicx}
\usepackage{amssymb}
\usepackage{amsmath}
\usepackage{esvect}
\usepackage{subfigure}
\usepackage{multirow}
\usepackage{tabularx}
\usepackage{makecell}
\usepackage{booktabs}
\usepackage{biblatex}  
\usepackage{marvosym}
\usepackage{bbm}

\usepackage{ragged2e}
\usepackage{pifont}
\usepackage{bm}
\usepackage{color}
\usepackage{tikz}
\usepackage{comment}
\usepackage{hyperref}
\hypersetup{colorlinks,
     citecolor=green,
     filecolor=green,
}
\usepackage{listings}
\definecolor{brightred}{RGB}{230, 0, 0} 
\definecolor{brightblue}{RGB}{0, 0, 230} 
\usepackage{verbatim}
\begin{document}
\title{Temporal Differential Fields for 4D Motion Modeling via Image-to-Video Synthesis}
\titlerunning{Mo-Diff}
%

\newcommand*\circled[1]{\tikz[baseline=(char.base)]{
            \node[shape=circle,draw,inner sep=1pt] (char) {#1};}}

\author{Xin You\inst{1, 2, 3, 4} \and
Minghui Zhang\inst{1, 2} \and
Hanxiao Zhang\inst{1} \and \\
Jie Yang\inst{1, 2(\text{\Letter})} \and
Nassir Navab\inst{3(\text{\Letter})}
}  
\authorrunning{Xin You et al.}
\institute{Institute of Medical Robotics, Shanghai Jiao Tong University, Shanghai, China \\
\and Institute of Image Processing and Pattern Recognition, Shanghai Jiaotong University, Shanghai, China \\
\and Computer Aided Medical Procedures,
Technical University of Munich, Munich, Germany \\
\and Munich Center for Machine Learning, Munich, Germany \\
\email{sjtu\_youxin@sjtu.edu.cn, nassir.navab@tum.de} \\
}

\maketitle              
\setlength{\textfloatsep}{6pt}
\begin{abstract}
Temporal modeling on regular respiration-induced motions is crucial to image-guided clinical applications. Existing methods cannot simulate temporal motions unless high-dose imaging scans including starting and ending frames exist simultaneously. However, in the preoperative data acquisition stage, the slight movement of patients may result in dynamic backgrounds between the first and last frames in a respiratory period. This additional deviation can hardly be removed by image registration, thus affecting the temporal modeling. To address that limitation, we pioneeringly simulate the regular motion process via the image-to-video (I2V) synthesis framework, which animates with the first frame to forecast future frames of a given length. Besides, to promote the temporal consistency of animated videos, we devise the Temporal Differential Diffusion Model to generate temporal differential fields, which measure the relative differential representations between adjacent frames. The prompt attention layer is devised for fine-grained differential fields, and the field augmented layer is adopted to better interact these fields with the I2V framework, promoting more accurate temporal variation of synthesized videos. Extensive results on ACDC cardiac and 4D Lung datasets reveal that our approach simulates 4D videos along the intrinsic motion trajectory, rivaling other competitive methods on perceptual similarity and temporal consistency. Codes are available at \url{https://github.com/AlexYouXin/Mo-Diff}


\end{abstract}
\section{Introduction}
4D Temporal modeling on breathing-induced motions, is significant to image-guided clinical applications \cite{ehrhardt20134d, jeung2012myocardial, wang2009dosimetric}, such as disease diagnosis and therapy planning. Particularly, 4D cardiac Magnetic Resonance Imaging (MRI), monitoring the anatomical variation of 3D cardiac structures \cite{amsalu2021spatial, you2023semantic}, is frequently used for cardiac disease analysis and surgical intervention. Besides, 4D pulmonary Computed Tomography (CT) is adopted to model the respiratory process, which is beneficial to the intraoperative puncture of pulmonary nodules by assisting clinicians to determine the optimal puncture path \cite{li2024robotic}.

Existing methods utilize flow-based interpolation models \cite{guo2020spatiotemporal, kim2024data} to deduce the regular motion process with the predefined starting and ending frames. And some researchers aim to synthesize the whole videos by means of diffusion models conditioning on these two prompting frames \cite{kim2022diffusion, ho2022video}. Basically speaking, all these methods could barely simulate breathing-induced motions only if two prompting frames exist simultaneously in the test-time evaluation. However, in the preoperative data acquisition stage, the slight movement or unstable breathing of patients may result in dynamic backgrounds between the first and last frames in a respiratory period. This additional bias cannot be thoroughly removed through image registration, thereby affecting the temporal motion modeling. Moreover, clinical research indicates that two high-dose CT or MRI scans will cause prolonged radiation exposure, potentially impairing patients' health \cite{power2016computed, you2024slord}. 

To address these limitations, we pioneeringly simulate regular respiration-induced motions via the image-to-video (I2V) synthesis framework, which animates with the first frame to forecast future frames of a given length. Owing to the compatibility of conditional diffusion models across different inputs \cite{zhang2023adding}, the image-to-video paradigm can be seamlessly plugged into diffusion models. Herein, the diffusion model implicitly learns temporal correlations of 4D data with regular motion discipline, under the conditional guidance of the starting frame and frame number. Besides, to promote the temporal consistency of animated videos, we propose the Temporal Differential Diffusion Model (TDDM) to generate temporal differential fields, which measure the relative differential representations between adjacent frames. The prompt attention layer (PAL) is devised for fine-grained differential fields, and the field augmented layer (FAL) is adopted to better interact these fields with the I2V framework, boosting precise temporal consistency of synthesized videos. Our two-stage pipeline is aimed at the \textbf{Mo}tion simulation via conditional \textbf{Diff}usion models, termed \textbf{Mo-Diff}. Extensive results on ACDC cardiac and 4D Lung datasets reveal that Mo-Diff forecasts accurate volumes along the intrinsic motion trajectory, rivaling other competitive methods on perceptual similarity and temporal consistency, which are more clinically significant than pixel-centric reconstruction metrics.

\noindent \textbf{Contributions:} 1) We pioneeringly introduce the image-to-video (I2V) paradigm to simulate regular breathing-induced motions, and propose the two-stage pipeline termed Mo-Diff. 2) The temporal differential diffusion model is devised to yield temporal differential fields as conditional guidance, amplifying the temporal consistency of synthesized volumes by I2V. 3) The prompt attention layer is aimed at fine-grained differential fields, and the field augmented layer is adopted to effectively interact these fields with the I2V framework, promoting more accurate temporal consistency. 4) Experiments demonstrate that Mo-Diff can synthesize precise and realistic videos with better perceptual quality and temporal consistency, showcasing potentials of simulating regular cardiac and pulmonary motions.


\noindent \textbf{Related work:} Breathing-induced motion modeling has been a critical research area within medical image computing and computer-assisted intervention \cite{wei2023mpvf, yuan20234d}. Existing methods on this topic can be categorized into two groups. The first group of approaches try to interpolate intermediate frames with the starting and ending frames in a breathing period, based on the perspective of optical flow estimation. Specifically, SVIN \cite{guo2020spatiotemporal} estimates forward and backward deformation fields, which will yield precise intermediate frames by linearly combining bidirectional information. UVI-Net \cite{kim2024data} utilizes the flow calculation model with the time-domain cycle-consistency constraint and linear motion hypothesis, to realize motion modeling in an unsupervised style.

Recently, denoising diffusion probabilistic models \cite{ho2020denoising, ho2022video, liu2024sora} show promising performance in generating realistic images or videos, by transforming random Gaussian distributions into target data distributions. The second category of methods maximize the potential of diffusion models to resolve video interpolation tasks in a generative manner. Essentially, these approaches implicitly learn the temporal motions from the starting frame to the ending frame \cite{jain2024video, bae2024conditional, chen2024ultrasound}. Furthermore, a recent work \cite{kim2022diffusion} devised a diffusion deformable model (DDM), which can learn spatial deformation maps between the starting and ending volumes and provide a latent code for generating intermediate frames along a geodesic path.

\begin{figure*}[!t]
\centerline{\includegraphics[width=0.98\linewidth]{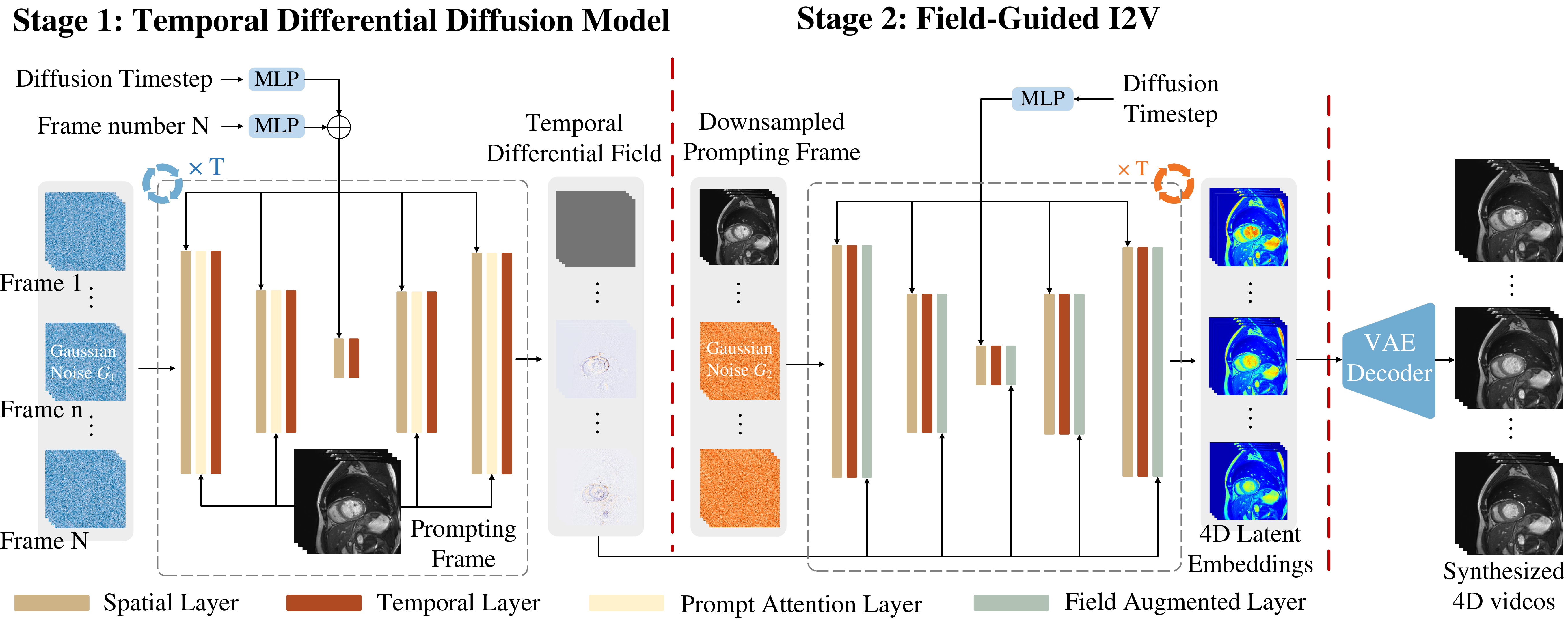}}
\caption{The whole Mo-Diff pipeline. Stage 1: TDDM yields temporal differential fields conditioning on the prompting frame and frame number. Stage 2: the conditional diffusion model will synthesize 4D latent embeddings with the guidance of temporal fields. These latent embeddings are then transformed into 4D videos with regular motions.}
\label{pipeline}
\end{figure*}

\section{Methodology}
\subsection{Image-to-Video Conditional Diffusion Models}
To eliminate the potential misalignment caused by patients' movements in clinical practice, we intend to avoid the data acquisition of the ending CT or MRI frame in a respiratory cycle, and only collect the starting frame for the motion simulation of future frames. Thus, we select the diffusion model conditioning on the first volume frame to simulate regular temporal motions. That generative model can also be viewed as the image-to-video (I2V) framework. For saving GPU memory, the image space is transformed into latent space with a well-trained Variational Autoencoder (VAE). And our framework draws lessons from the Latent Video Diffusion Model (LVDM) \cite{blattmann2023align, rombach2022high}. It conducts the denoising process in the latent space. Before training, the input video $x_{0}$ is first
encoded into a latent embedding $z_{0}$ = $\mathcal{E}(x_{0})$ with VAE
encoder $\mathcal{E}(\cdot)$, and $z_{0}$ can be restored into $x_{0}$ via VAE decoder $\mathcal{D}(\cdot)$. 
Then the latent code $z_{0}$ is perturbed as:
\begin{eqnarray}
  & z_{t} = \sqrt{\overline{\alpha}_{t}} \; z_{0} + \sqrt{1 - \overline{\alpha}_{t}} \; \epsilon,  \epsilon \sim \bm{N}(0, 1)
\end{eqnarray}
where $\overline{\alpha}_{t} = \prod_{i=1}^{t} (1 - \beta_{t}) $ with $\beta_{t}$ is the noise strength coefficient
at time step $t$, and $t$ is uniformly sampled from the timestep index set
${1, . . . , T}$. This process can be regarded as a Markov chain, which
incrementally adds Gaussian noise to the latent code $z_{0}$. The denoising model $\epsilon_{\theta}$ receives $z_{t}$ as input and is optimized to learn the
latent space distribution with the objective function:
\begin{eqnarray}
  & \mathcal{L}_{\epsilon} = E_{z_{t}, \epsilon \sim \mathcal{N}(0, 1)} \; || \epsilon - \epsilon_{\theta} (z_{t}, t, c) ||^{2}
\label{diffusion loss}
\end{eqnarray}
where $c$ represents the condition, and $\epsilon_{\theta}$ is implemented as a light-weight U-Net. Using the devised LVDM, we aim to synthesize the latent embeddings showcasing motion dynamics of anatomical structures.


\subsection{Temporal Differential Diffusion Model}
The I2V model is aimed at modeling the motion discipline of the breathing process. However, as demonstrated by \cite{liu2024sora, shi2024motion}, synthesized videos are prone to visualizing sequences with poor temporal consistency if no extra condition is injected into diffusion models. Thus, we devise the temporal differential diffusion model (TDDM) to generate temporal differential fields as an additional prompt. 
This conditional input can boost a temporally consistent synthesis of latent embeddings by the above-mentioned I2V framework, with smoother variation on anatomical shape and texture details.

Temporal differential fields are defined to measure the relative variance between two adjacent frames. 
Specifically, the $i$th frame of temporal fields $\mathcal{F}_{i}$ is calculated with the subtraction between image frame $\mathcal{I}_{i}$ and $\mathcal{I}_{i-1}$, with $i$ ranging from $1$ to frame number $N$. The detailed process is noted as follows:
\begin{eqnarray}
    \mathcal{F}_{i} = 
\begin{cases}
    \mathcal{I}_{i} \ominus \mathcal{I}_{i-1}, & \text{if } i \geq 2 \\
0, & \text{otherwise }
\end{cases}
\end{eqnarray}
$\mathcal{F}_{1}$ is set as a zero mask to improve the convergence efficiency. 
After the score-based training, TDDM can transform the Gaussian noise $\mathcal{G}_{1} \sim \mathcal{N}(0, 1)$ into synthesized differential fields $\hat{\mathcal{F}}$ corresponding to the prompting image $\mathcal{I}_{1}$ and frame number $N$, formulated as $\hat{\mathcal{F}} = TDDM(\mathcal{G}_{1}, t; \mathcal{I}_{1}, N)$. Here $t$ refers to the diffusion time step. As revealed in Fig. \ref{pipeline}, $\hat{\mathcal{F}}$ will suppress dynamically varying backgrounds, and highlight the foreground regions in the respiratory process.
Thus, $\hat{\mathcal{F}}$ can enormously boost the reconstruction performance of future frames by the I2V network.
Besides, temporal fields representing regular motions will significantly improve the temporal consistency of synthesized videos.

It is worth mentioning that frame number $N$ is a significant conditional variable, serving as a compensation for the absence of the ending frame. Due to a fixed samping duration between frames, the specific $N$ corresponds to the specific breathing period, acquired by electrocardiogram signals, thus influencing the rate of temporal motion variations. As shown in Fig. \ref{pipeline}, $N$ serves as an explicit constraint on the motion velocity of respiration, promoting TDDM to generate appropriate temporal differential fields, further improving the video synthesis with regular motions.

\subsection{Network Details of Mo-Diff}
\textbf{Spatial \& Temporal Layer:} Mo-Diff is designed for yielding four-dimensional videos, which pose a huge challenge to the joint modeling on spatial and temporal domains. Inspired by \cite{qiu2017learning}, we try to extract spatial and temporal information in a separate manner. Given the noisy input $\mathcal{I}_{\mathcal{G}}$ with a dimension of $B \times C \times N \times L \times H \times W$ (corresponding to batch size, channel, frame number, length, height, and width individually), we respectively implement 3D spatial convolutions and 1D temporal convolutions on $\mathcal{I}_{\mathcal{G}}$ as visualized by Fig. \ref{pipeline}(b).

\begin{figure*}[!t]
\centerline{\includegraphics[width=0.90\linewidth]{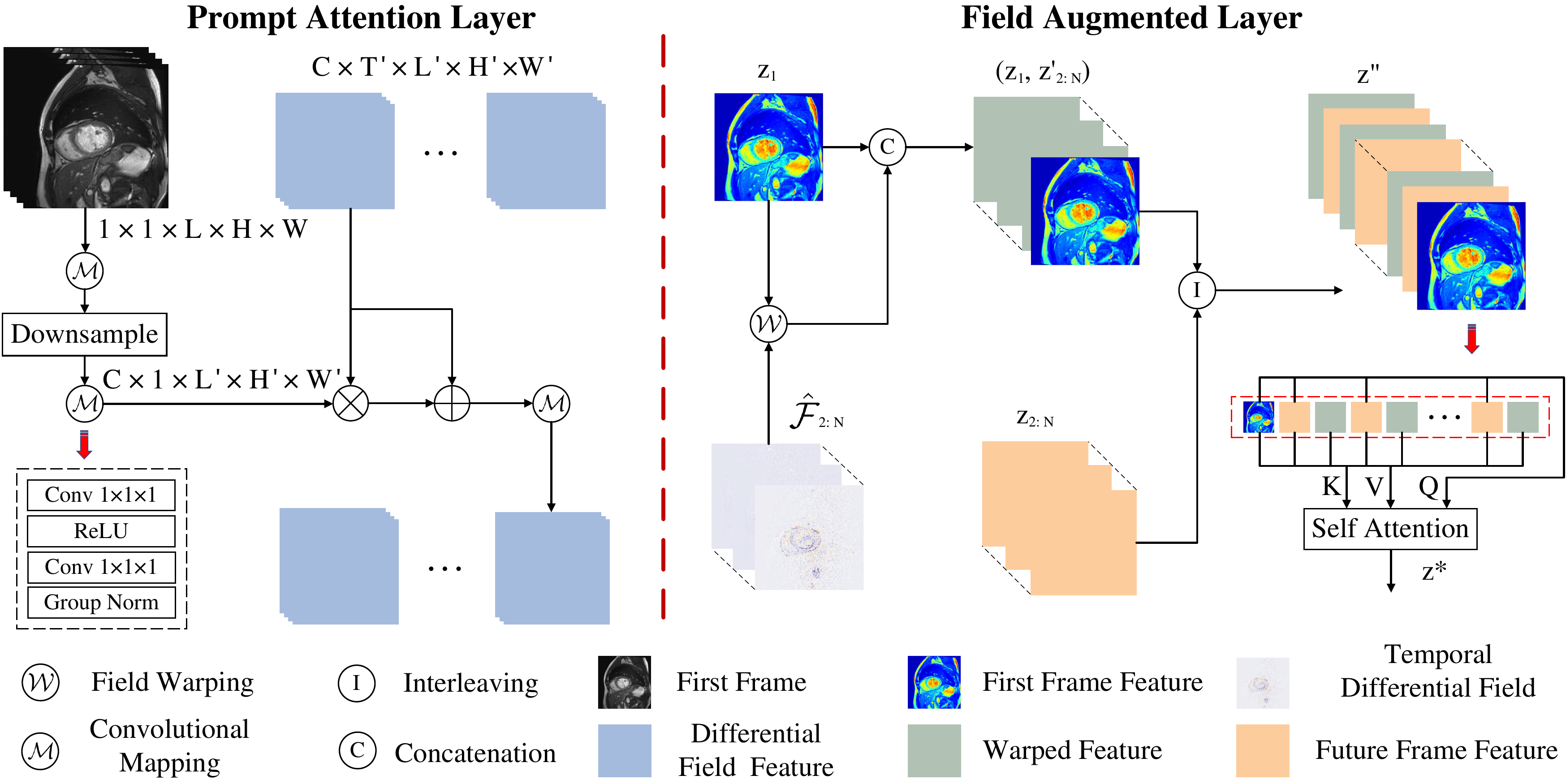}}
\caption{The detailed structure of prompt attention layer and field augmented layer.}
\label{PAL}
\end{figure*}

\noindent \textbf{Prompt Attention Layer (PAL):} To integrate the prompt information into the first-stage TDDM, we do not simply concatenate the first frame with noisy differential fields for training as what previous works did \cite{xing2024cross, peng2023generating, zhang2024pass}. Instead, a feature-level fusion is adopted to synthesize temporal differential fields with fine-grained textures and shapes, which hold importance for the synthesis performance of future frames. Thus, we propose the PAL to boost the representation of TDDM. Specifically, given the initial frame $\mathcal{I}_{1}$, we define the convolutional mapping $\mathcal{M}$ which can transform the image space into feature space. As visualized in Fig. \ref{PAL}, $\mathcal{M}$ is a combination of convolution, ReLU, and group normalization operations. Then the warped feature of the prompting frame is utilized to enhance features corresponding to temporal differential fields via the element-wise multiplication and addition.

\noindent \textbf{Field Augmented Layer (FAL):} To synthesize 4D videos with temporal consistency, we propose the FAL to interact temporal differential fields with frame features. Specifically, $z_{1}$ denotes the feature of the first frame, and $z_{2:N}$ means the subsequent frame features. As revealed in Fig. \ref{PAL}, via the warping transform $\mathcal{W}$, differential fields of future frames $\hat{\mathcal{F}}_{2:N}$ are warped into frame features $z'_{2:N}$. The detailed process can be formulated as Eq \eqref{warping}.
\begin{eqnarray}
    & z'_{i} = \mathcal{W}(z_{1}, \hat{\mathcal{F}}_{1 \rightarrow i}), \, \, i = 2, . . . , N 
  \label{warping}
\end{eqnarray}
$z'_{i}$ shows similar spatial-temporal patterns to $z_{1}$. Then warped features are interleaved with future frame features $z_{2:N}$ in the temporal dimension. Specifically, two-branch features with the same frame index are arranged side by side, and the merged result is noted as augmented features $z''=[z_{1}, z_{2}, z'_{2}, ... , z_{N}, z'_{N}]$, which is then projected as key and value vectors. And the query vector arises from $z$. After the self-attention operation, augmented frame features $z_{\*}$ can be attained to synthesize temporal-consistent latent embeddings.

\begin{table}[!t]
  \begin{center}
    \caption{Baseline comparison with various models. Double prompting frame: starting and ending frames; Single prompting frame: only the starting frame (Blue, red values represent best evaluation metrics corresponding to two types of models respectively).}
\label{benchmark_table}
    \centering
    \resizebox{0.95\textwidth}{!}{
    \begin{tabular}{cccccccc}
        \hline
                \hline
  \multirow{2}{*}{Prompting}  & \multirow{2}*{Model} & \multicolumn{3}{c}{ACDC Cardiac } & \multicolumn{3}{c}{4D Lung}  \\  
  \cmidrule(r){3-5}    \cmidrule(r){6-8}
  Frame & & PSNR (dB) $\uparrow$ & LPIPS $\downarrow$ & FVD $\downarrow$ & PSNR (dB) $\uparrow$ & LPIPS $\downarrow$ & FVD $\downarrow$ \\
            \hline
     &    SVIN \cite{guo2020spatiotemporal} & 31.43\scriptsize{$\pm0.421$} & \textcolor{brightblue}{1.563\scriptsize{$\pm0.206$}} & \textcolor{brightblue}{93.6} & 30.49\scriptsize{$\pm0.304$} & 2.650\scriptsize{$\pm0.245$} & 125.6 \\ 
     &    Voxelmorph \cite{balakrishnan2019voxelmorph} & 30.77\scriptsize{$\pm0.502$} & 1.969\scriptsize{$\pm0.197$} & 102.1 & 29.90\scriptsize{$\pm0.373$} & 2.815\scriptsize{$\pm0.260$} & 149.0 \\
  Double   &   UVI-Net \cite{kim2024data} & \textcolor{brightblue}{32.16\scriptsize{$\pm0.402$}} & 1.662\scriptsize{$\pm0.245$} & 94.2 & \textcolor{brightblue}{31.57\scriptsize{$\pm0.311$}} & \textcolor{brightblue}{2.211\scriptsize{$\pm0.216$}} & \textcolor{brightblue}{121.7}  \\ 
     &    LDMVFI \cite{danier2024ldmvfi}  & 27.11\scriptsize{$\pm0.460$} & 2.943\scriptsize{$\pm0.410$} & 99.2 & 26.31\scriptsize{$\pm0.453$} & 3.659\scriptsize{$\pm0.341$} & 142.7 \\
     &    DDM \cite{kim2022diffusion} & 29.79\scriptsize{$\pm0.504$} & 2.689\scriptsize{$\pm0.352$} & 110.3 & 29.67\scriptsize{$\pm0.420$} & 2.905\scriptsize{$\pm0.330$} & 165.5 \\
            \hline
     &      LDDM \cite{chen2024ultrasound} & 24.53\scriptsize{$\pm0.481$} & 2.634\scriptsize{$\pm0.308$} & 105.7 & 25.19\scriptsize{$\pm0.273$} & 2.914\scriptsize{$\pm0.287$} &  146.3 \\
  Single  &     Condi-Diffusion \cite{ho2022video} & 26.59\scriptsize{$\pm0.545$} & 2.460\scriptsize{$\pm0.357$} & 95.7 & 25.95\scriptsize{$\pm0.391$} & 3.294\scriptsize{$\pm0.375$} & 133.0 \\ 
    &       Mo-Diff  & \textcolor{brightred}{30.79\scriptsize{$\pm0.409$}} & \textcolor{brightred}{1.317\scriptsize{$\pm0.189$}} & \textcolor{brightred}{86.1} & \textcolor{brightred}{29.86\scriptsize{$\pm0.282$}} & \textcolor{brightred}{2.137\scriptsize{$\pm0.225$}} & \textcolor{brightred}{115.8} \\
        \hline 
            \hline
    \end{tabular}}
  \end{center}
\end{table}


\section{Experiment}
\subsection{Experimental Settings}
\textbf{Dataset.} To evaluate the efficacy of Mo-Diff, we conduct experiments on public ACDC cardiac \cite{bernard2018deep} and 4D-Lung datasets \cite{hugo2016data}. For ACDC with the MRI modality, volume sequences from the diastolic to systolic phases are extracted as 4D data. Of all the 150 4D sequences, cases with identity $1\mbox{-}100$, $101\mbox{-}120$, and $121\mbox{-}150$ serve as the training, validation, and testing sets. Besides, all MRI volumes are resampled with a voxel space of $1.5 \times 1.5 \times 3.12mm^{3}$ and cropped to $128 \times 128 \times 32$ \cite{you2024learning}. And the frame number $N$ shows a range of $[6, 16]$. For 4D-Lung CT images, the end-inspiratory and end-expiratory scans are set as the initial and final images. We collect 125 4D videos from \cite{hugo2016data}, with resolutions equal to $10 \times 128 \times 128 \times 128$. They are split as $80/15/30$ cases for training, validation, and inference. The intensity value is scaled as $[0, 1]$. We choose quantitative evaluation metrics including PSNR \cite{hore2010image}, Learned Perceptual Image Patch Similarity (LPIPS) \cite{zhang2018unreasonable}, and Fréchet Video Distance (FVD) \cite{unterthiner2018towards}. PSNR is the traditional pixel-centric reconstruction metric, which cannot well assess the quality of synthesized videos in clinical practice \cite{jain2024video}. Instead, 
LPIPS and FVD are more reasonable metrics representing human visions. Specifically, LPIPS reveals perceptual variance. And FVD evaluates the feature distribution bias between ground truths and synthesized videos, measuring temporal consistency and continuity.

\noindent \textbf{Implementation Details.} The VAE maps the image space into the downsampled latent space with a ratio of $1/4$. For the proposed TDDM, the temporal channel of inputs is set as $16$, but we only calculate the score-based loss $\mathcal{L}_{\epsilon}$ with the first N channels for efficient training. And the synthesized 4D latent embeddings by field-guided I2V are further transformed into temporal sequences via the latent decoder of VAE. All models are trained using AdamW optimizer with the linear warm-up strategy. The initial learning rate is set as $1e\mbox{-}4$ with a cosine learning rate decay scheduler, and weight decay is set as $1e\mbox{-}5$. The batch size, diffusion step $T$, and training epoch are equal to $2$, $1000$, and $500$. Experiments are implemented based on Pytorch and 2 NVIDIA RTX 4090 GPUs.

\begin{figure*}[!t]
\centerline{\includegraphics[width=0.95\linewidth]{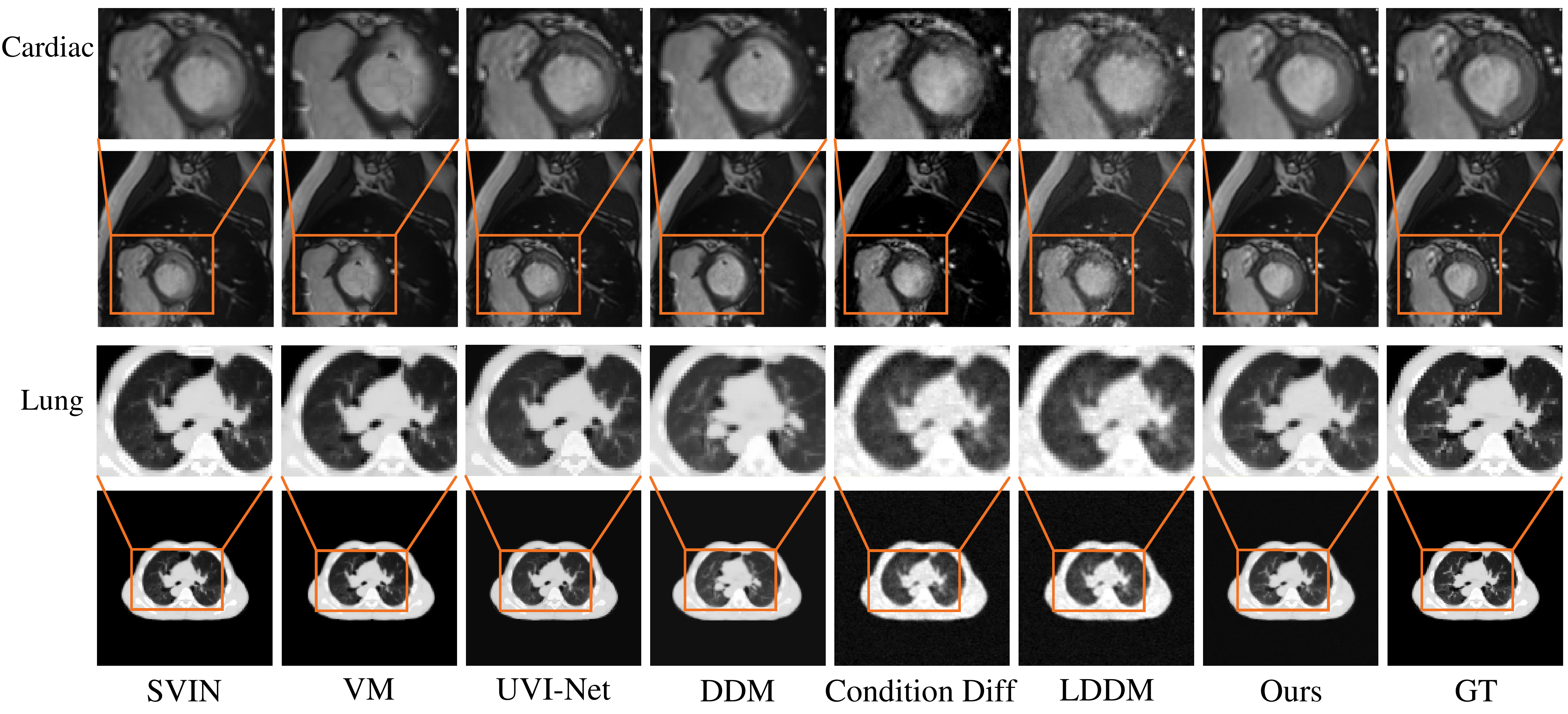}}
\caption{Qualitative comparison of intermediate frames between different models.}
\label{FIGURE_BENCHMARK}
\end{figure*}

\subsection{Experimental Results}
\textbf{Baseline Comparison.} For performance comparisons on the simulation of 4D temporal motions, we list out two categories of methods, classified as $"$Double$"$ (conditioning on both starting and ending frames) and $"$Single$"$ (conditioning on only the starting frame). Obviously, models with two prompting frames can achieve better pixel-wise reconstruction performance, with higher PSNR values. However, when evaluating the feature-level perceptual similarity with human judgments, our proposed model reveals a good LPIPS metric on ACDC, with a $0.246\downarrow$ lower value than supervision-based model SVIN \cite{guo2020spatiotemporal} as illustrated by Table \ref{ablation_table}. Also, Mo-Diff synthesizes 4D videos with more consistent temporal distributions with ground truth videos, outperforming UVI-Net \cite{kim2024data} with a $5.9\downarrow$ FVD value. That convincingly validates the efficacy of the TDDM.

\begin{figure*}[!t]
\centerline{\includegraphics[width=0.96\linewidth]{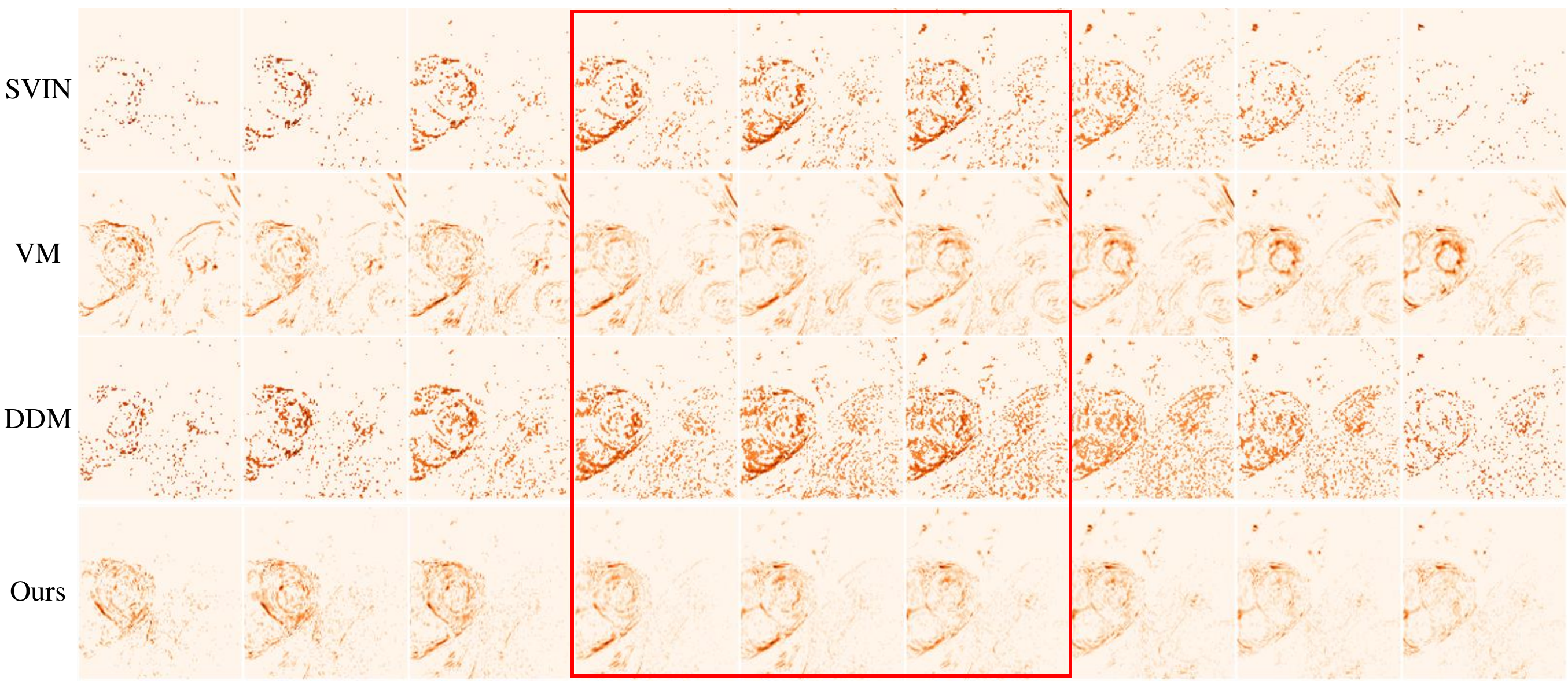}}
\caption{Temporal error maps between different models. All frame predictions are visualized except starting and ending frames (Red box: intermediate frames).}
\label{Temporal_error_map}
\end{figure*}

\noindent \textbf{Qualitative Results.} As illustrated by Fig. \ref{FIGURE_BENCHMARK}, our proposed model can forecast precise enough frames, rivaling other competitive flow-based models. Specifically, Mo-Diff reveals a promising shape synthesis of arteries for lung data and texture synthesis of myocardium for cardiac data. Also, Fig. \ref{Temporal_error_map} depicts temporal error maps for a qualitative evaluation on temporal consistency. Our results show more precise simulation for three intermediate frames, which are tricky to forecast due to a farther distance to prompting frames. In contrast, SVIN and Voxelmorph reveal better reconstruction performance for bilateral frames.

\begin{table}[!t]
  \begin{center}
  \caption{Ablation study on key components. w/o PAL: replaced with a channel concatenation operation; w/o FAL:  concatenate channels between frame features and differential fields, then conduct 1D temporal self-attention.}
\label{ablation_table}
  \resizebox{0.80\textwidth}{!}{
  \begin{tabular}{ccccccccc}  
  \hline  
  \multirow{2}*{\scriptsize{Frame}} & \multirow{2}*{\scriptsize{PAL}} & \multirow{2}*{\scriptsize{FAL}} & \multicolumn{3}{c}{ACDC Cardiac} & \multicolumn{3}{c}{4D Lung}  \\  
  \cmidrule(r){4-6}    \cmidrule(r){7-9}
  \scriptsize{Number} & & & PSNR (dB) $\uparrow$ & LPIPS $\downarrow$ & FVD $\downarrow$ & PSNR (dB) $\uparrow$ & LPIPS $\downarrow$ & FVD $\downarrow$ \\
  \hline   
  \textcolor{red}{\ding{55}}  & \textcolor{green}{\ding{52}} & \textcolor{green}{\ding{52}} & 28.07 & 2.206 & 109.7 & 27.44 & 2.732 & 142.9 \\
  \textcolor{green}{\ding{52}}  & \textcolor{red}{\ding{55}} & \textcolor{red}{\ding{55}} & 28.60 & 1.878 & 111.5 & 27.93 & 2.560 & 131.3 \\
  \textcolor{green}{\ding{52}}   & \textcolor{green}{\ding{52}} & \textcolor{red}{\ding{55}} & 29.86 & 1.683 & 94.7 & 28.46 & 2.351 & 119.4  \\
  \textcolor{green}{\ding{52}}  & \textcolor{red}{\ding{55}} & \textcolor{green}{\ding{52}} & 29.97 & 1.532 & 100.9 & 29.12 & 2.215 & 128.0 \\
  \textcolor{green}{\ding{52}}  & \textcolor{green}{\ding{52}} & \textcolor{green}{\ding{52}} & \textbf{30.79} & \textbf{1.317} & \textbf{86.1} & \textbf{29.86} & \textbf{2.137} & \textbf{115.8}  \\
  \hline 
  \end{tabular}}
  \end{center}
\end{table}

\subsection{Ablation Study}
We carried out the ablation study on Mo-Diff. 1) \textbf{Frame number $N$}: as shown in Table \ref{ablation_table}, removing the frame number off Mo-Diff will largely degrade the reconstruction and synthesis performance, with $2.42dB\downarrow$ PSNR and $0.889\uparrow$ LPIPS on ACDC. Without $N$, the model cannot perceive the implicit period information of a respiratory process, affecting the motion modeling. 2) \textbf{PAL}: PAL aims to fully leverage the first frame by extracting prompting frame features. A coarse channel concatenation fails to effectively inject prompting information into TDDM, resulting in unsatisfactory temporal differential fields. 3) \textbf{FAL}: replacing FAL with a channel fusion between these fields and frame features, will lead to modality confusion, which further affects synthesis performance with $1.40dB\downarrow$ PSNR and $0.214\uparrow$ LPIPS.

\section{Conclusion}
We pioneeringly simulate regular respiration motions via the proposed Mo-Diff framework, which animates with the first
frame to forecast future frames with a given length. Besides, the temporal
differential diffusion model can generate temporal differential fields to promote the temporal consistency of animated videos. Mo-Diff can model relatively regular temporal motions via the prompting frame. However, it requires more clinical guidance, including electrocardiogram signals, to simulate highly unstable breathing, which is promising in future work.

\printbibliography
\end{document}